\documentclass[preprint,12pt]{elsarticle}

\usepackage{amssymb}

\usepackage{array} 
\usepackage{booktabs} 
\usepackage{caption} 

\usepackage{xcolor} 
\usepackage[T1]{fontenc}
\usepackage{booktabs}
\usepackage{multirow}
\usepackage[utf8]{inputenc}
\usepackage[T1]{fontenc}

\journal{}

\begin{document}

\begin{frontmatter}



\title{Amplifying Pathological Detection in EEG Signaling Pathways through Cross-Dataset Transfer Learning}



\author[mila,FL,udem,*]{Mohammad-Javad Darvishi-Bayazi}
\author[nrc]{Mohammad Sajjad Ghaemi}
\author[mila,udem]{Timothee Lesort}
\author[mila,udem]{Md Rifat Arefin}
\author[FL,udem]{Jocelyn Faubert}
\author[mila,udem]{Irina Rish}

\affiliation[mila]{organization={Mila},
            addressline={Québec AI Institute}, 
            city={Montréal},
            state={QC},
            country={Canada}}
\affiliation[FL]{organization={Faubert Lab},
            city={Montréal},
            state={QC},
            country={Canada}}
\affiliation[udem]{organization={Université de Montréal},
            city={Montréal},
            state={QC},
            country={Canada}}
\affiliation[nrc]{organization={National Research Council Canada},
            city={Toronto},
            state={ON},
            country={Canada}}
\affiliation[*]{organization={Corresponding author},
                email={mohammad.bayazi@mila.quebec}
                }

\begin{abstract}
Pathology diagnosis based on EEG signals and decoding brain activity holds immense importance in understanding neurological disorders. With the advancement of artificial intelligence methods and machine learning techniques, the potential for accurate data-driven diagnoses and effective treatments has grown significantly. However, applying machine learning algorithms to real-world datasets presents diverse challenges at multiple levels. The scarcity of labelled data, especially in low regime scenarios with limited availability of real patient cohorts due to high costs of recruitment, underscores the vital deployment of scaling and transfer learning techniques. In this study, we explore a real-world pathology classification task to highlight the effectiveness of data and model scaling and cross-dataset knowledge transfer. As such, we observe varying performance improvements through data scaling, indicating the need for careful evaluation and labelling. Additionally, we identify the challenges of possible \textit{negative transfer} and emphasize the significance of some key components to overcome distribution shifts and potential spurious correlations and achieve positive transfer. 
We see improvement in the performance of the target model on the target (NMT) datasets by using the knowledge from the source dataset (TUAB) when a low amount of labelled data was available.
Our findings indicate a small and generic model (e.g. ShallowNet) performs well on a single dataset, however, a larger model (e.g. TCN) performs better on transfer and learning from a larger and diverse dataset. 


\end{abstract}






\end{frontmatter}


\section{Introduction}
\label{Introduction}
In recent years, the field of pathology diagnosis has witnessed significant advancements through the utilization of electroencephalogram (EEG) signals and the decoding of brain activity \cite{albaqami2023automatic, zhong2023automated}. These techniques offer valuable insights into the functioning of the human brain and provide critical information for understanding and diagnosing various neurological disorders. Machine learning approaches have played a pivotal role in analyzing EEG data, enabling the development of automated systems for pathology classification.

However, despite the remarkable progress made, the application of machine learning algorithms to real-world datasets presents a set of challenges \cite{roy2019deep}. One significant challenge arises from the scarcity of labelled data, particularly in scenarios where the number of instances is limited, commonly referred to as low regime data. In such cases, conventional machine learning techniques may struggle to achieve satisfactory performance due to the insufficient amount of data available for accurate model training \cite{adadi2021survey}.

To address this data scarcity issue, efforts have been made to create large-scale datasets by recording EEG signals directly from patients in hospitals \cite{khan2022nmt, obeid2016temple}. These initiatives aim to gather diverse and comprehensive data, encompassing a wide range of neurological conditions and patient demographics. By collecting data in real-world clinical settings, these datasets capture the complexity and heterogeneity of pathology cases encountered in actual practice.

Another challenge that arises from applying machine learning algorithms to real-world EEG datasets is the low generalization performance due to distribution shift \cite{lan2018domain} and noise \cite{he2021data}. Distribution shift refers to the phenomenon where the data distribution of the test set differs from that of the training set, which can lead to a degradation of the model's accuracy on unseen data \cite{gulrajani2020search}. Noise refers to the unwanted variations or distortions in the EEG signals, which can be caused by various factors, such as electrode placement, muscle artifacts, eye movements, or environmental interference \cite{roy2019deep}. These factors can affect the quality and reliability of the EEG data, making it difficult for machine learning models to extract meaningful features and patterns \cite{banville2022robust}.

One possible solution to overcome these challenges is to use transfer learning, which is a technique that allows a model that has been trained on a source task or domain to be adapted to a new target task or domain by transferring some of its knowledge \cite{wan2021review}. Transfer learning can leverage the information learned from a large and diverse source dataset to improve the performance of a model on a small and specific target dataset. 
Transfer learning significantly reduces the training time and enhances accuracy. Furthermore, this approach can effectively address the challenges of data insufficiency, variability, and inefficiency that are common in brain-computer interface applications \cite{yang2023cross}.
However, transfer learning also poses some risks, such as negative transfer, which occurs when the transferred knowledge is irrelevant or harmful to the target task or domain \cite{zhang2022survey}. Negative transfer can result in a worse performance than training from scratch, and it can be caused by various factors, such as domain discrepancy, task discrepancy, or label discrepancy \cite{tran2022plex}. Therefore, it is important to select an appropriate source dataset and a suitable transfer learning method for the target task or domain.

In this work, we investigated the knowledge transferability across two pathological detection datasets, The widely used Temple University Hospital Abnormal (TUAB) \cite{obeid2016temple} and the new NUST-MH-TUKL (NMT) scalp EEG dataset \cite{khan2022nmt}, which contain thousands of recordings labelled as normal or abnormal by a team of qualified neurologists in different population and hospitals. We studied how and when a transfer of gained knowledge is possible. Furthermore, we shed light on the black box complex structure of deep neural networks via the Centered Kernel Alignment (CKA) similarity method. CKA metric evaluates the degree of similarity between the representations learned by different neural networks \cite{kornblith2019similarity}. Our work improved the performance of the target model by leveraging knowledge from the source dataset (TUAB) to the target (NMT) datasets.


\section{Materials and Method} \label{Materials&Method}

\subsection{Datasets}

\begin{figure}[th]
\centering 
\includegraphics[width=0.95\textwidth]{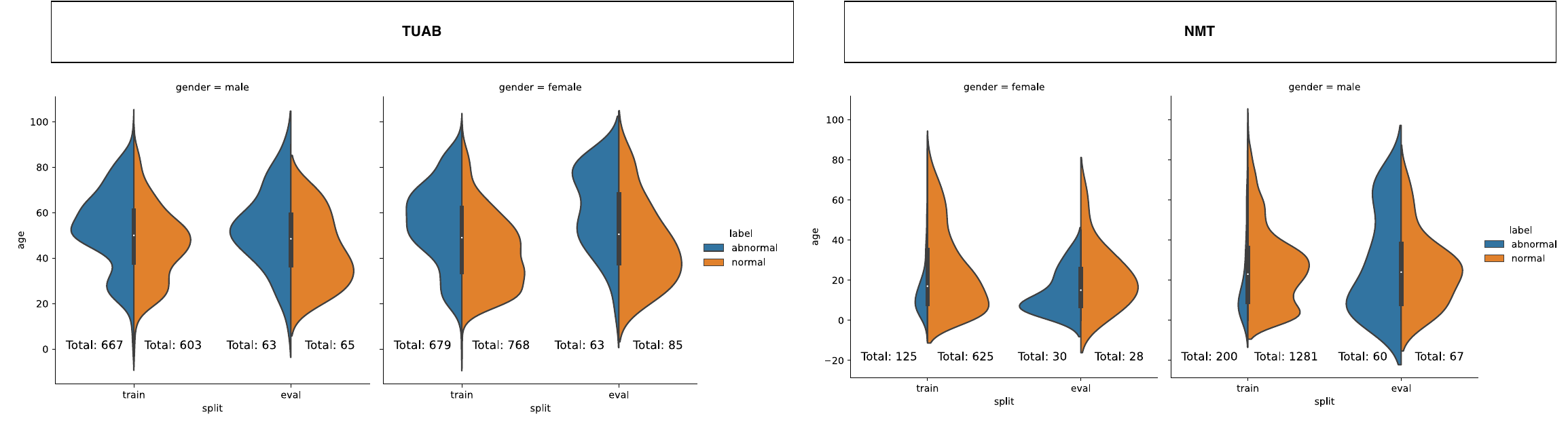} 
\caption{A comparison of the data distribution with respect to age in two EEG datasets: TUAB and NMT. The figure shows that the statistics between splits, gender and datasets vary drastically. The age distribution differs in data splits for gender and pathological states. The number of samples in different conditions is not balanced, which may affect the performance of the models.} 
\label{fig:data_des}
\end{figure}
Our experiment is based on the TUAB Abnormal EEG Corpus3 (v3.0.0) \cite{obeid2016temple}, which is a publicly available dataset of significant size that facilitates the task of decoding general pathology from EEGs. The corpus comprises 2993 recordings, each lasting at least 15 minutes, obtained from 2329 unique patients. It is divided into a training and a final test set. The corpus includes recordings of both male and female patients from a broad age range, spanning from 7 days to 96 years, covering infants, children, adolescents, adults, and seniors. The dataset includes various pathologies diagnosed in patients, such as epilepsy, strokes, depression, and Alzheimer's disease, among others. However, only binary labels are available.

The NMT EEG dataset is an open-source annotated dataset of healthy and pathological EEG recordings \cite{khan2022nmt}. The dataset consists of 2,417 recordings from unique participants. Each recording is labelled as either normal or abnormal by a team of expert neurologists. The dataset also includes demographic information such as the gender and age of the patients. The dataset focuses on the South Asian population and can be used for training machine learning models to identify normal and abnormal EEG patterns. The dataset is intended to increase the diversity of EEG datasets and to overcome the scarcity of accurately annotated publicly available datasets for EEG research.

The dataset is imbalanced in pathological and gender classes, in the training and statistics shift in the test sets, which may pose a challenge for machine learning algorithms. In the training set, there are 66.38\% male and 33.62\% female subjects, with 13.50\% of the male subjects and 16.67\% of the female subjects having abnormal EEG recordings. In the test set, there are 68.65\% male and 31.35\% female subjects, with 47.24\% of the male subjects and 51.72\% of the female subjects having abnormal EEG recordings.

Figure \ref{fig:data_des} illustrates the challenge of generalization and transfer across different subsets and datasets, as the distribution of groups changes drastically in each case. For example, the NMT dataset has a lower proportion of abnormal recordings in the training than the TUAB dataset. Moreover, within each dataset, the train and evaluation subsets have different class balances, which can affect the performance and robustness of the models. Therefore, it is important to consider these factors when designing and evaluating the models for EEG analysis. 
Each dataset has a validation set, which we use for testing the models. We did not use any information from this set during training, to avoid data leakage. We split the training sets into 85\% and 15\% for training and validation, respectively. The validation set was used to select the best model based on the balanced accuracy metric. The test set was used to evaluate the final performance of the selected model on unseen data.

For the Preprocessing step, We applied the same preprocessing steps to both datasets, which are TUAB and Temple. These steps are: We selected a subset of 21 electrode positions following the international 10–20 placement because these electrode positions were common to all the recordings in the datasets. We discarded the first 60 seconds of every recording, We used a maximum of 20 minutes of every recording, to avoid considerable feature generation and resampling times for exceptionally long recordings.
We downsampled the EEG recordings to 100 Hz and clipped them at $\pm 800 \mu V$ to reject unphysiologically extreme values and to ensure comparability to previous works \cite{kiessner2023extended,khan2022nmt}. 
We z-scored each channel of the recordings based on the channel mean and standard deviation, to normalize the data and reduce the effect of outliers.
We also normalized each channel in each trial to range [0 1], to make the data more suitable for the neural network models.

\subsection{Models}
To evaluate the performance of different models, we trained several well-known EEG models while altering the number of recordings in the training set. It is worth noting that the test set remained fixed to the original split of the dataset. We experimented with five models with at least five different seeds for each of them. These experiments allowed us to thoroughly evaluate the effectiveness of data size and parameter size and select the best-performing model for decoding general pathology from EEGs. We used all models without making any changes to their original structure.

The four models that we experimented with are:
\begin{itemize}
    \item EEGNet \cite{lawhern2018eegnet}: EEGNet is a compact convolutional neural network that can be used for EEG-based brain-computer interfaces. It is designed to work with different EEG paradigms, such as P300, ERN, MRCP, and SMR, and to be as efficient as possible. It has only three convolutional layers and 2018 trainable parameters, which makes it easy to train and deploy. It uses batch normalization and ELU activation to improve its performance and robustness.
    \item ShallowNet \cite{schirrmeister2017deep}: A shallow convolutional neural network that consists of two convolutional layers, each followed by a max-pooling layer and a dropout layer. Both layers use batch normalization and ELU activation. The final layer is a softmax layer that outputs the class probabilities. ShallowNet has 36722 trainable parameters.
    \item Deep4Net \cite{schirrmeister2017deep}: A deep convolutional neural network that consists of four convolutional blocks, each followed by a max-pooling layer and a dropout layer. The convolutional blocks have different numbers of filters, kernel sizes, and strides, and use batch normalization and ELU activation. The final layer is a softmax layer that outputs the class probabilities. Deep4Net has 277052 trainable parameters and was originally proposed for EEG-based emotion recognition. 
    \item TCN-EEG \cite{elsken2019neural}: A temporal convolutional network that consists of several residual blocks, each containing two dilated causal convolutional layers and different dilation rates. The residual blocks use weight normalization and ReLU activation. The final layer is a softmax layer that outputs the class probabilities. TCN-EEG has 456502 trainable parameters.
\end{itemize}

Table \ref{tab:models} shows the summary of the models that we used in our experiments. It includes the name, the architecture, the number of parameters, and the original reference of each model. The models are Deep4Net, ShallowNet, TCN-EEG and EEGNet. They are all convolutional neural networks that can be applied to EEG signal analysis. However, they differ in their structure, complexity, and specificity. Some of them are more generic and adaptable, while others are more tailored and efficient for specific EEG tasks or domains.

\begin{table}[th]
\centering
\caption{Models details } 
\begin{tabular}{c|c|c}
\toprule
\hline 
Model Name & \# of parameters & Type \\
\hline
TCN \cite{elsken2019neural}& 456502 & Temporal ConvNet \\
Deep4Net \cite{schirrmeister2017deep} & 277052 & 4-layer ConvNet \\
ShallowNet \cite{schirrmeister2017deep} & 36722 & 1-layer ConvNet \\
EEGNet \cite{lawhern2018eegnet} & 2018 & compact ConvNet  \\
\hline 
\bottomrule
\end{tabular} 
\label{tab:models}
\end{table}

\subsection{Training}
We followed the cropped training approach proposed by \cite{schirrmeister2017deep}, which divides the EEG signals into equally sized, maximally overlapping segments that match the receptive field of the networks. The receptive field is the number of signal samples that the networks can process at a time, which varies depending on the network architecture. All networks, except the TCN, had a receptive field of approximately 600 samples, while the TCN had a receptive field of approximately 900 samples. We used the AdamW optimizer \cite{loshchilov2017decoupled} to optimize the categorical cross-entropy loss function, which measures the difference between the predicted and true class probabilities. AdamW is an improved version of Adam \cite{kingma2014adam} that decouples the weight decay updates and the loss function optimization, which leads to better generalization \cite{loshchilov2017decoupled}. We used cosine annealing \cite{loshchilov2016sgdr} to schedule the learning rates for both the gradient and weight decay updates. Cosine annealing is a technique that gradually reduces the learning rates according to a cosine function, which helps to avoid local minima and find better solutions. We did not perform learning rate restarts, which are optional steps that reset the learning rates to their initial values at certain intervals. To have an unbiased selection of hyper-parameters, We used the hyper-parameters used in this work \cite{kiessner2023extended} (See Section \ref{sec:HPs} for more details).

\subsection{Data Augmentation}
We used data augmentation for EEG based on this previous work \cite{rommel2022data}, which proposed and compared different methods of augmenting EEG data to improve the performance and robustness of machine learning models. We used four methods of data augmentation, each with a probability of 0.1:

\begin{itemize}
    \item SignFlip: This method randomly flips the sign of the EEG signals, which simulates a change in the polarity of the electrodes.
    \item ChannelsDropout: This method randomly drops out some channels of the EEG signals, which simulates a loss of contact or a malfunction of the electrodes. We used a dropout probability of 0.2.
    \item FrequencyShift: This method randomly shifts the frequency spectrum of the EEG signals, which simulates a change in the sampling rate or drift in the frequency band. We used a maximum frequency shift of 2 Hz.
    \item SmoothTimeMask: This method randomly masks some time segments of the EEG signals with a smooth transition, which simulates a temporary occlusion or a distortion of the signals. We used a mask length of 600 samples.
    \item BandstopFilter: This method randomly applies a band-stop filter to the EEG signals, which removes a narrow frequency band from the signals. This simulates a noise reduction or a notch filter. We used a bandwidth of 1 Hz.
    \item ChannelsShuffle: This method randomly shuffles the order of the channels of the EEG signals, which changes the spatial configuration of the signals. This simulates a different electrode placement or a permutation of the channels. We used a shuffle probability of 0.2.
\end{itemize}

\subsection{Transfer Learning}
Transfer learning is a technique that allows a model that has been trained on a source task or domain to be adapted to a new target task or domain by transferring some of its knowledge \cite{fawaz2018transfer}. Fine-tuning is a common method of transfer learning, which involves adjusting the parameters of the model on the target data while keeping the initial weights from the source data. Transfer learning and fine-tuning can help to improve the performance and generalization of the model, especially when the target data is limited and similar to the source data \cite{wan2021review}.

In this work, we applied transfer learning and fine-tuning to our EEG models, which were trained on two different datasets: TUAB and NMT. TUAB is a large and diverse dataset of EEG recordings from normal and abnormal subjects, while NMT is a smaller and more specific dataset of EEG recordings from normal and pathological subjects. We used TUAB as the source dataset and NMT as the target dataset, and we experimented with four models: Deep4Net, ShallowNet, TCN-EEG and  EEGNet.

We trained the models on the whole source dataset (TUAB) and used the weights of that model to initialize the network to train on the second dataset (NMT). This way, we hoped to leverage the knowledge learned from TUAB to improve the performance on NMT. To have a baseline, we also trained the same models with random initialization on the target dataset (NMT), without using any information from TUAB. We compared these two conditions (fine-tuned the pre-trained model vs. randomly initialized model) on the test set of NMT, using balanced accuracy as our evaluation metric.

\subsection{Representation similarity}
\citet{cianfarani2022understanding} provides a novel perspective on how robust training methods affect the internal representations learned by neural networks. The authors use representation similarity metrics, such as centred kernel alignment (CKA) \cite{kornblith2019similarity}, to compare the representations of different layers of robust and non-robust networks on three vision datasets. They reveal some interesting properties of robust representations, such as less specialization, more homogeneity and higher overfitting. Inspired by this work, we experimented with different scenarios to assess the transferability between these two datasets and models trained on them. We wanted to investigate how the similarity of representations influences the ability of a model to adapt to a new domain or task.

\section{Results} \label{Results}
\subsection{Transfer learning}
\begin{figure}[th]
\centering \includegraphics[width=0.99\textwidth]{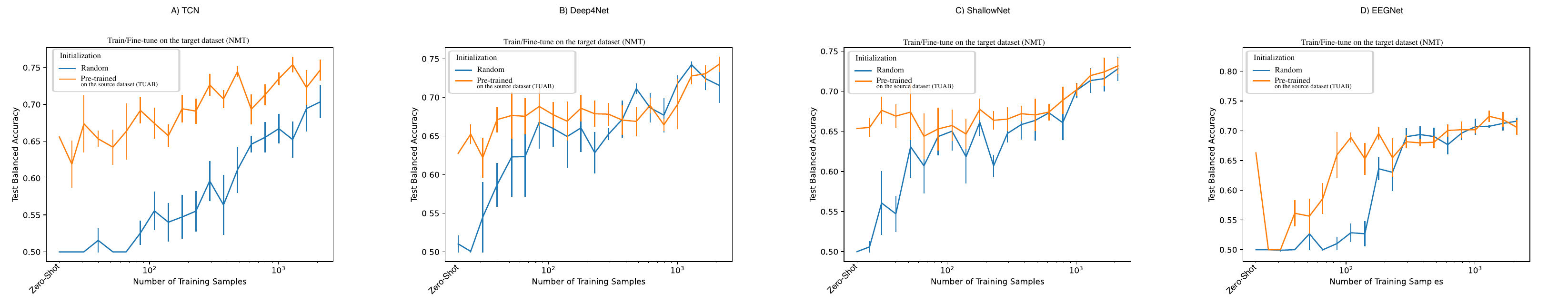} 
\caption{The effect of pretraining on pathology detection balanced accuracy in the NMT dataset. The x-axis shows the number of training samples and the y-axis shows the test balanced-accuracy. The orange line represents the performance of fine-tuning a pretrained model and the blue line represents the baseline model trained from scratch. The error bars show the standard error of the mean across five random seeds.} \label{fig:pretraining_effect}
\end{figure}

The results in Figure \ref{fig:pretraining_effect} indicate that pretraining enhances pathology detection performance, particularly in the low data regime (<100 samples). This is crucial due to the low sample size in clinical studies, which poses a challenge for deep learning approaches. Our results suggest that the transfer is feasible if several criteria are satisfied, such as data normalization and adequate preprocessing.


\begin{figure}[th]
\centering \includegraphics[width=0.99\textwidth]{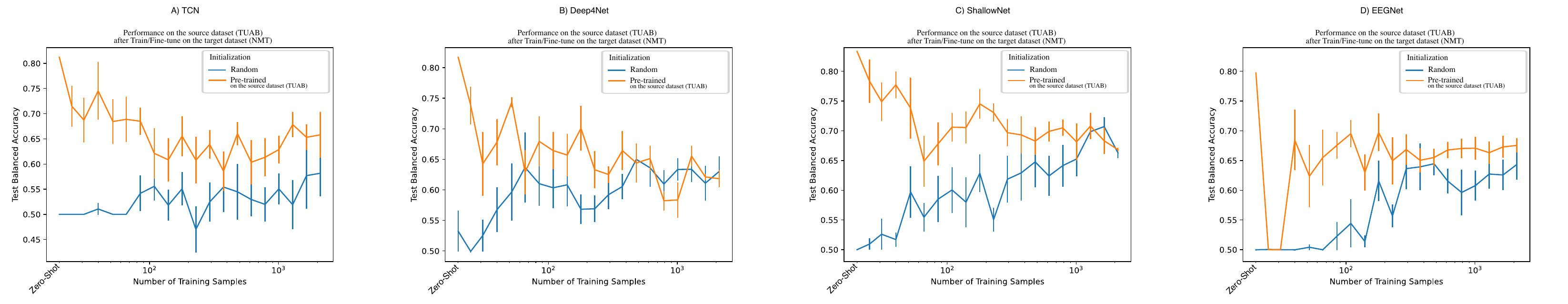} 
\caption{The performance on the source dataset (TUAB) while finetuning or training on the target dataset (NMT). The x-axis shows the number of training samples from the target dataset and the y-axis shows the balanced accuracy on the source dataset. The orange line represents the pretrained model that is finetuned on the target dataset and the blue line represents the baseline model that is trained from scratch on the target dataset. 
} \label{fig:cat_forgetting}
\end{figure}

The results in Figure \ref{fig:cat_forgetting} show that the pretrained model does not maintain high performance on the source dataset while adapting to the target dataset. The performance of the baseline model on the source dataset increases a bit as it trains on the target dataset. The results also show a sign of catastrophic forgetting in these real-world datasets. 

To enable a model to work on both datasets, we opt to train a model on the concatenate and train on both datasets as suggested in \cite{zhang2022survey}. However, this approach faces the challenges of class imbalance and dataset bias, which affect the model’s performance and generalization. To address these challenges, we designed a data sampler that samples uniformly from each dataset and each class, ensuring a balanced and fair representation of the data. The performance of the resulting network is reported as follows:

\begin{table}[th]
\caption{Performance of models on different distributions}
\label{tab:merge}
\begin{minipage}{\linewidth}
\def\arraystretch{0.9}\tabcolsep=1.5pt\small\centering
\begin{tabular}{@{}lllll@{}}
\toprule
Model                     & Train on                    & BAC TUAB                & BAC NMT                   & TUAB+NMT                \\ \midrule
\multirow{4}{*}{TCN}      & TUAB                        & 0.8169$\pm$0.0035          & \textbf{0.6304 $\pm$ 0.0200} & 0.7471$\pm$0.0062          \\ \cmidrule(l){2-5} 
                          & NMT                         & 0.5816$\pm$0.0990          & 0.7034$\pm$0.0479            & 0.6342$\pm$0.0748          \\ \cmidrule(l){2-5} 
                          & TUAB+NMT                    & 0.8092$\pm$0.0073          & \textbf{0.7353$\pm$0.0192}   & \textbf{0.7804$\pm$0.0119} \\ \cmidrule(l){2-5} 
                          & PtTUAB\textgreater FtNMT \footnote{Pre-train a model on TUAB and Fine-tuned on NMT} & 0.6592$\pm$0.0441          & \textbf{0.7370$\pm$0.0332}   & 0.6935$\pm$0.0250          \\ \midrule
\multirow{4}{*}{Deep4Net} & TUAB                        & 0.8164$\pm$0.0052          & 0.6277 $\pm$ 0.0163          & 0.7482$\pm$0.0081          \\ \cmidrule(l){2-5} 
                          & NMT                         & 0.6300$\pm$0.0540          & 0.7156$\pm$0.0487            & 0.6688$\pm$0.0517          \\ \cmidrule(l){2-5} 
                          & TUAB+NMT                    & 0.8090$\pm$0.0112          & 0.7295$\pm$0.0134            & 0.7751$\pm$0.0077          \\ \cmidrule(l){2-5} 
                          & PtTUAB\textgreater FtNMT & 0.5900$\pm$0.0506          & 0.7334$\pm$0.0332            & 0.6518$\pm$0.0381          \\ \midrule
\multirow{4}{*}{ShallowNet} & TUAB & \textbf{0.8240$\pm$0.0074} & 0.6202$\pm$0.0201 & \textbf{0.7496$\pm$0.0118} \\ \cmidrule(l){2-5} 
                          & NMT                         & \textbf{0.6627$\pm$0.0160} & \textbf{0.7281$\pm$0.0285}   & \textbf{0.6922$\pm$0.0092} \\ \cmidrule(l){2-5} 
                          & TUAB+NMT                    & \textbf{0.8098$\pm$0.0114} & 0.7152$\pm$0.0123            & 0.7709$\pm$0.0055          \\ \cmidrule(l){2-5} 
                          & PtTUAB\textgreater FtNMT & \textbf{0.6682$\pm$0.0037} & 0.7318$\pm$0.0161            & 0.6973$\pm$0.0073          \\ \midrule
\multirow{4}{*}{EEGNet}   & TUAB                        & 0.8140$\pm$0.006           & 0.6145$\pm$0.0422            & 0.7419$\pm$0.0173          \\ \cmidrule(l){2-5} 
                          & NMT                         & 0.6209$\pm$0.0657          & 0.7063$\pm$0.0236            & 0.6602$\pm$0.0451          \\ \cmidrule(l){2-5} 
                          & TUAB+NMT                    & 0.8090$\pm$0.0128          & 0.7152$\pm$0.0123            & 0.7585$\pm$0.0152          \\ \cmidrule(l){2-5} 
                          & PtTUAB\textgreater FtNMT & 0.6582$\pm$0.0097          & 0.6971$\pm$0.0298            & 0.6782$\pm$0.0175          \\ \bottomrule
\end{tabular}
\end{minipage}
\end{table}

Table \ref{tab:merge} shows the performance of the models under different training and test distributions. We can observe that the models trained on TUAB, which is a large and diverse dataset, have a high balanced accuracy (BAC) on TUAB itself, but a low BAC on NMT, which is a small and noisy dataset. This indicates that the models suffer from distribution shifts and overfitting when they encounter data from a different domain. For example, when we train on TUAB, the average performance on the TUAB dataset is 0.8178 and it drops by 0.1946 to 0.6232 when we test on NMT, which is out of its distribution. On the other hand, the BAC on NMT when trained on NMT is 0.71335, which we assume is in distribution accuracy. So the difference between a classifier that has not seen any data from the NMT dataset and one that has seen the whole NMT data is 0.0901. This shows that there is still room for improvement in transferring knowledge from TUAB to NMT.

Table \ref{tab:merge} also shows the performance of models that have been trained on the concatenation of both datasets, TUAB and NMT. We can see that these models have comparable BAC on both datasets compared to models trained on a single dataset. It was expected that combining data from different domains can improve the robustness and generalization of the models, as they can learn from a larger and more diverse set of examples. However, these models have a lower BAC than the models fine-tuned with our proposed framework, which indicates that simply concatenating data is not enough to achieve optimal transfer learning. We also see that as the model is bigger, it can get better BAC on the concatenated data, since the model has more capacity. For example, TCN, which has the largest number of parameters among the four architectures, has the highest BAC on both TUAB and NMT when trained on the concatenated data and ShallowNet which is a small and simple network that has the best accuracies in single datasets training. This implies that larger models can benefit more from combining data from different domains, as they can learn more complex and diverse features.


\subsection{Representation similarity}\label{Representation}

\begin{figure}[th]
\centering 
\includegraphics[width=0.75\textwidth]{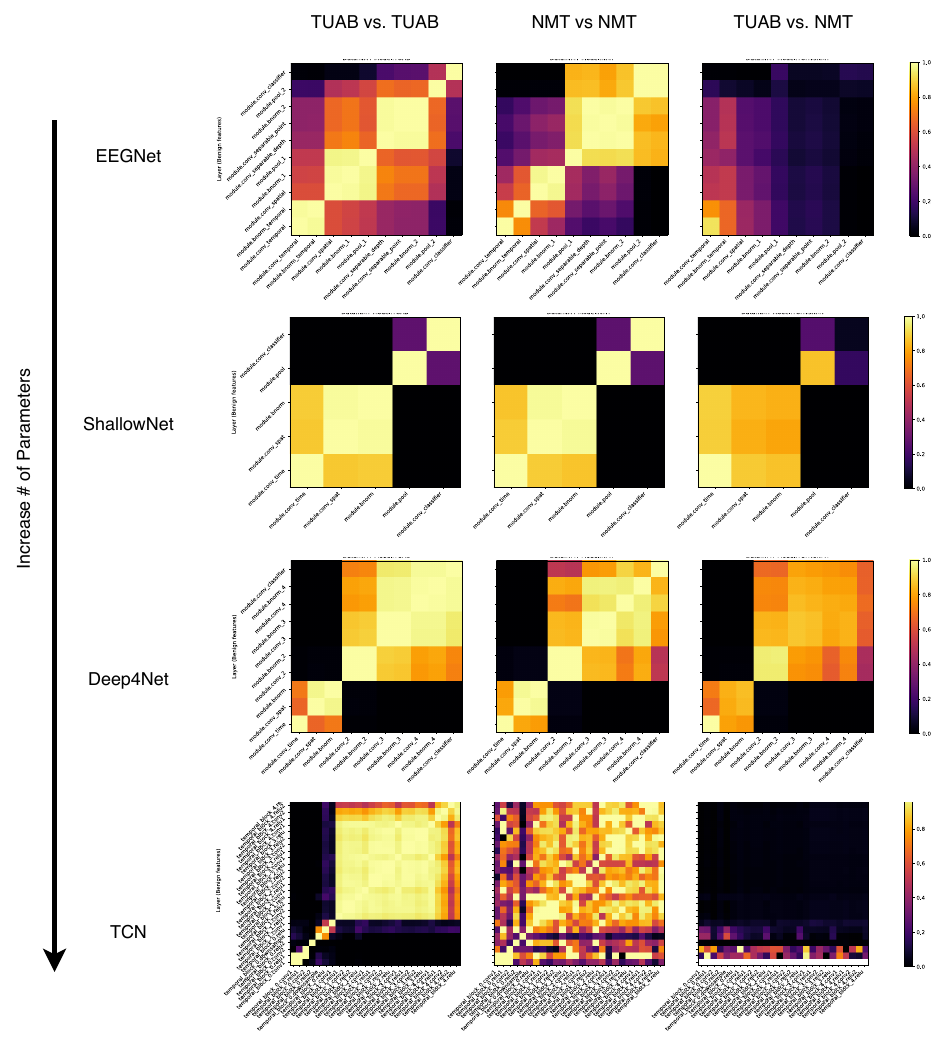} 
\caption{The similarity between different layers of networks. From the left, the first and second are within a model and the last one is cross-model similarity.}
\label{fig:r_sim}
\end{figure}

Figure \ref{fig:r_sim} shows the degree of similarity between different layers of networks trained on TUAB and NMT datasets. The networks trained on TUAB exhibit more block structure as the network size increases, indicating that the layers are more specialized and less redundant. The networks trained on NMT, on the other hand, have more uniform similarity across layers, suggesting that the layers are more general and complementary. The similarity between TUAB and NMT networks decreases from the early layers to the last layers, as shown in Figure \ref{fig:r_sim} (right), implying that the datasets have different representations at higher levels of abstraction.

\subsection{Discriminative Fine-Tuning }
\begin{figure}[th]
\centering \includegraphics[width=0.75\textwidth]{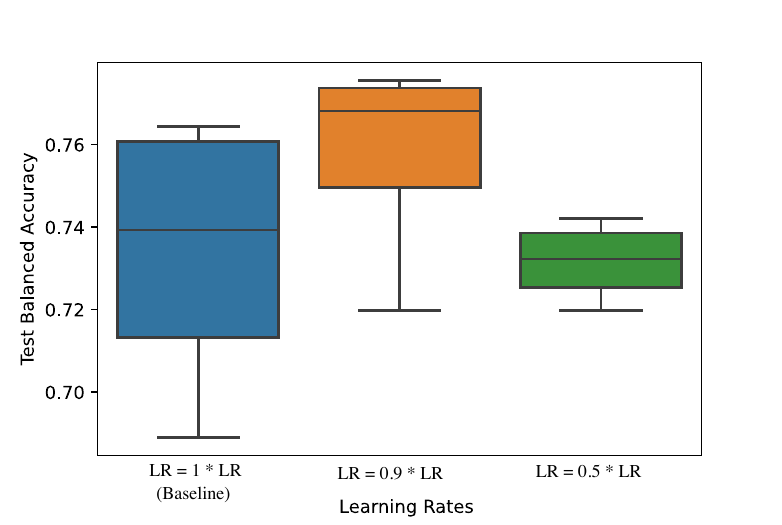} 
\caption{
The effect of discriminative fine-tuning on the balanced accuracy (BAC) of the model for the NMT dataset. The x-axis shows Discriminative Learning Rates, and the y-axis shows the BAC score. The figure shows that the model with discriminative fine-tuning achieves a higher BAC score than the baseline model, indicating that it can better adapt to the new domain and task.
} 
\label{fig:dis_ft}
\end{figure}

Based on insights in Section \ref{Representation}, we hypothesize if we adjust the learning rate during the fine-tuning process from the last layer to the early layers, giving more weight to the lower layers that are more similar across datasets and less weight to the higher layers that are more divergent would improve the transfer performance. In the literature, it is called Discriminative Fine-Tuning.
Discriminative Fine-Tuning is a technique for transferring knowledge from a pretrained model to a downstream task. It involves using different learning rates for different layers of the model, depending on how relevant they are to the task. The idea is to fine-tune the lower layers more slowly since they capture general features, and the higher layers more quickly since they capture task-specific features \cite{howard2018universal}. It has been shown that it works for MRI fine-tuning \cite{malik2022youtube}. Here as our results confirm these assumptions, we fine-tune the model using the discriminative fine-tuning technique. This allows us to leverage the pretrained knowledge from the TUAB dataset and adapt it to the NMT dataset, which has a smaller size and a different domain. We compare the performance of our model with different learning rate schedules and report the results in Figure \ref{fig:dis_ft}. 

Figure \ref{fig:dis_ft} shows the impact of different learning rate decay factors on the performance of the model. The learning rate decay factor is the ratio between the learning rates of two consecutive layers. The x-axis shows the learning rate decay factor, and the y-axis shows the balanced accuracy (BAC) score. The figure shows that the model with discriminative fine-tuning achieves a higher BAC score than the baseline model when the learning rate decay factor is 0.9, indicating that it can better adapt to the new domain and task. However, when the learning rate decay factor is 0.5, the performance of the model with discriminative fine-tuning drops significantly, suggesting that it is too aggressive and causes instability in the training process. Therefore, choosing an appropriate learning rate decay factor is important for discriminative fine-tuning, as it adds a new hyperparameter to the model.

\subsection{Pre-Training (Why Deep neural networks?)}\label{TUAB_scaling}

\begin{figure}[th]
\centering \includegraphics[width=0.85\textwidth]{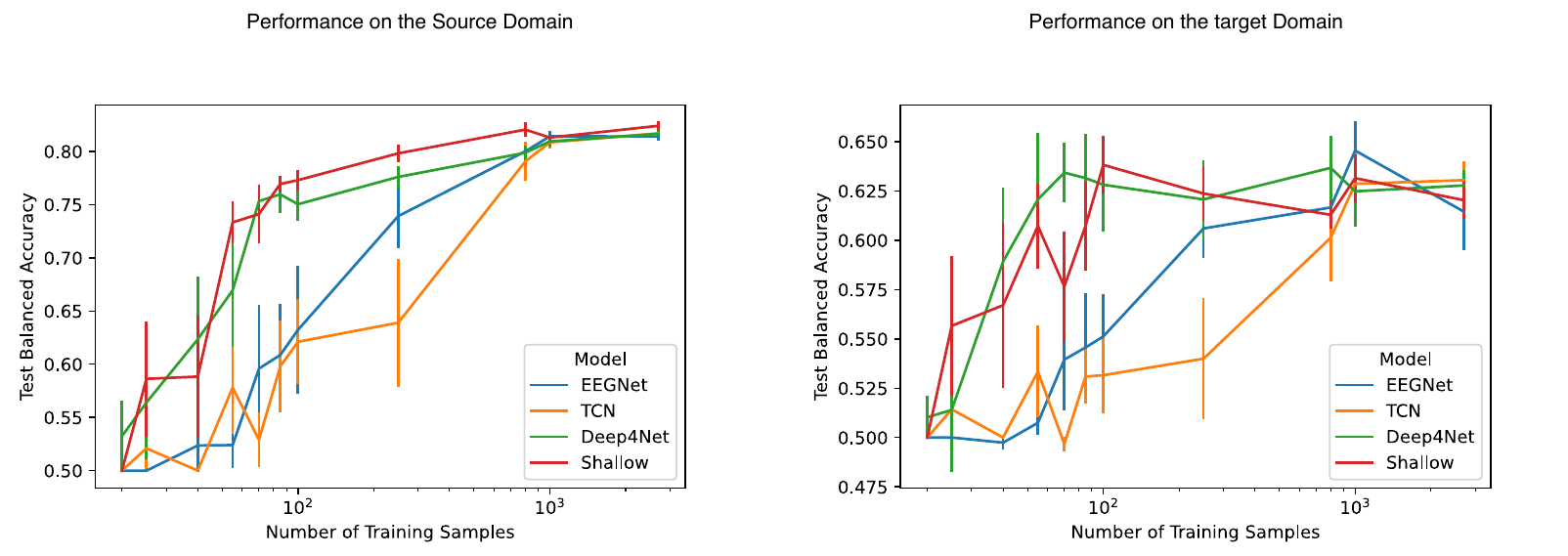} 
\caption{Scaling behaviour of different EEG models on the TUAB (Source) and NMT (Target) dataset. It illustrates the balanced accuracy plotted against the number of recordings in the training set for various models. The accuracy increases as the number of recordings increases, but the curve is saturated after reaching 1000 recordings. The error bars represent the standard error.
} \label{fig:TUAB_scaling}
\end{figure}

After analyzing the results in Figure \ref{fig:TUAB_scaling}, we observed that all the models performed similarly after $500$ for the number of recordings, with varying slopes. All models show a saturation accuracy of around 80\%.

Deep4Net and ShallowNet are two deep learning architectures that can be used to classify EEG signals into different mental states. They are more generic architectures than some other models that are specifically designed for EEG analysis, such as EEGNet or TCN. This means that they have fewer inductive biases or assumptions about the structure and distribution of the EEG data. This can be an advantage when dealing with diverse and heterogeneous EEG data from different experiments and subjects. Both Deep4Net and ShallowNet show a good performance for different available recordings in the source and target domain, meaning that they can adapt well to new EEG tasks or domains. This indicates that they can learn the essential features of EEG signals that are invariant across different scenarios. However, Deep4Net has more layers and parameters than ShallowNet, which gives it more capacity to adapt to the new task in the fine-tuning stage. In fact, it was shown in the previous work \cite{khan2022nmt} that Deep4Net has positive transfer, meaning that it improves its performance after fine-tuning, while ShallowNet has negative transfer, meaning that it degrades its performance after fine-tuning. Therefore, we chose Deep4Net for fine-tuning our model for the NMT dataset.

\section{Discussion} \label{Discussion}

In figure \ref{fig:TUAB_scaling} all models show a saturation accuracy of around 80\% in the source task (TUAB) which may have arisen from label noise or inter-rater agreement, as discussed in previous literature \cite{gemein2020machine,kiessner2023extended}. Based on our findings, we can conclude that approximately 500 recordings are sufficient for this task, and adding more recordings beyond 500 until the total of 2700 did not significantly improve the performance of the models. These results suggest that adding more data beyond a certain point may not always yield better results, and careful consideration of the amount of data used in training is essential for optimal model performance. Furthermore, we also identified that there are 54 common subjects in the two pathological groups (normal/abnormal) in the TUAB dataset. There are 149 out of 2993 rows that have their pathological status changed, some of them within a few days. This could be one of the possible sources of label noise that affects the model performance and generalization. This also might be due to noisy input data as it is a problem in other medical domains (e.g. magnetic resonance imaging \cite{schulz2020different}), which can be mitigated by transfer learning \cite{malik2022youtube}.

\citet{gemein2020machine} developed a feature-based decoding framework for analyzing EEG data and demonstrates that it achieves accuracies similar to state-of-the-art deep neural networks. The accuracies obtained ranged from 81 to 86\% across both approaches, and analysis showed that they used similar aspects of the data, such as delta and theta band power at temporal electrode locations. The study argues that the accuracies of current binary EEG pathology decoders could saturate near 90\% due to imperfect inter-rater agreement of clinical labels and that such decoders are already clinically useful where experts are rare. Our results confirm these findings, meaning that we also observe a saturation accuracy of around 80\% for all the networks when trained on the TUAB dataset. However, we also notice that although all the networks achieve similar accuracy when trained on all available recordings, their scaling and transfer behaviour are different. More generic networks, such as Deep4Net and ShallowNet, perform better than more specific networks, such as EEGNet and TCN for the amount of data. This means that they improve their performance more rapidly and consistently as more data is added. Moreover, more generic networks also transfer better than more specific networks when fine-tuned on a new task or domain. 

The difference between the reported accuracy in our results and the previous results comes from the fact that we chose the best model based on a validation set and report results on the predefined test set, while they might have used Oracle model selection \cite{gulrajani2020search}. Oracle model selection is a method of choosing the best model based on the test set performance, which can lead to overfitting and optimistic results. By using a validation set, we avoid this problem and ensure that our results are more realistic and generalizable. We also report balanced accuracy (BAcc) as our evaluation metric, which is the average of the recall obtained in each class. BAcc is a suitable metric for ML when the goal is to minimize the overall classification error, especially when the classes are imbalanced \cite{tholke2023class}.

It's worth noting that there have been several efforts to extend clinical EEG datasets with a larger number of recordings. For instance, a recent study proposed a dataset \cite{kiessner2023extended} with 15,300 automatically labelled EEG recordings and evaluated and improved the performance of machine learning algorithms on this dataset. However, our results in Section \ref{TUAB_scaling} suggest that for this specific task of general pathology decoding from EEGs, we do not need more than 500 recordings to achieve good model performance. This may be due to the richness and diversity of the TUAB Abnormal EEG Corpus3 (v3.0.0), which contains recordings from patients of various ages and with different pathologies. However, more recording in the source domain might improve the pretrained model. Further research could investigate whether this conclusion holds for other types of EEG analysis tasks or if more data is necessary for different contexts. Our findings suggest that we might be in the early stages of the broken scaling laws \cite{caballero2022broken} where improvements are minimal, but significant improvements could be achieved beyond a certain point, which is consistent with the non-monotonic transitions observed in neural scaling behaviour.



In practical situations, any automated screening algorithm is highly likely to encounter data from sources it has not seen before. Therefore, it is crucial to assess how well different architectures can handle variations in the devices and sources used for data acquisition. Ideally, deep learning algorithms should consistently perform well across different datasets. However, this is not always the case. Evaluating the performance of abnormal EEG detection across datasets has been difficult due to the lack of publicly available datasets specific to this problem. To address this, In \cite{khan2022nmt} they tested the generalization performance of several CNN-based architectures. They trained them on the TUAB dataset and then evaluated their performance on the NMT dataset. They reported that there was a noticeable decline in performance, with the Shallow CNN architecture achieving an accuracy of 45\% and an AUC of 0.48, and the Deep CNN architecture achieving an accuracy of 48\% and an AUC of 0.46. Similar performance degradation was observed when these architectures were trained on the NMT dataset and tested on the TUAB dataset. These results emphasize the importance of collecting diverse datasets from multiple sources, as algorithms trained on data from a single source struggle to generalize well in the case of EEG data. 
In contrast to the challenges mentioned earlier, our findings provide encouraging evidence that the two datasets, TUAB and NMT, exhibit transferability and perform well in the context of our study. Despite the anticipated performance degradation when applying the Deep and Shallow CNN architectures trained on one dataset to the other, we observed that these models still achieved satisfactory results. Specifically, the Deep4Net and Shallow CNN architecture attained the best accuracy of $63\% \pm 0.05$ when tested on the NMT dataset, and the performance degradation was 17\%.  These findings suggest that, to some extent, the models trained on one dataset were able to generalize to the other dataset without fine-tuning.


The results in Figure \ref{fig:cat_forgetting} indicate that the models tend to forget the previous knowledge when they are fine-tuned on a new task or domain. This is a common problem in ML known as catastrophic forgetting, which limits the ability of the models to learn from sequential or diverse data. This problem poses a challenge and an opportunity for the fields of continual learning and out-of-distribution generalization, which aim to develop methods and techniques that can enable the models to learn continuously from new data without forgetting the old ones and to generalize well to unseen or novel data that differ from the training data. These fields can benefit from exploring these real-world datasets, which can provide realistic and complex scenarios for testing and improving their methods and techniques.

\section{Conclusion and outlook}\label{Conclusion}
In this work, we presented an approach for EEG pathology detection based on transfer learning.
Our methodology has the potential to improve the transferability performance of neural network models with noisy and small sample size datasets by leveraging knowledge from large and diverse datasets.
We conducted extensive experiments on two real-world EEG datasets, TUAB and NMT. We provide insights into the transferability of neural network features for EEG pathology detection by analyzing and visualizing the CKA similarity scores of the feature maps.

Our work has several implications and contributions to the field of EEG pathology detection and transfer learning. 
We demonstrated that transfer learning emerges as a potent methodology to tackle the challenge of data scarcity issue that hampers the application of machine learning algorithms to real-world medical datasets.
We applied CKA as a novel and effective measure of similarity between neural network representations of EEG signals, that can serve as a guiding tool for the transfer learning procedure and for elucidating the interpretability and explainability of the learned features. We proposed a simple yet highly efficient method to identify the optimal transfer point through the CKA similarity scores. This method has a huge potential to enhance model generalization and robustness significantly. We provided insights into the scaling and transfer aspects of the networks, that enable researchers and practitioners in choosing the optimal network architecture and data size for their EEG analysis tasks. 

\newpage
\section*{Acknowledgements}
This work was funded by Canada CIFAR AI Chair Program and from the Canada Excellence Research Chairs (CERC) program, National Research Council Canada, Natural Sciences and Engineering Research Council (NSERC-CAE-CRIAC-CARIQ, NSERC discovery grant RGPIN-2022-05122), Doctoral Research Microsoft Diversity Award (Microsoft-Mila), Faculty of medicine-UdeM, and Faculté des études supérieures et postdoctorales. We thank Compute Canada for providing computational resources.

\newpage
\appendix

\section{Hyper-parameters} \label{sec:HPs}

\begin{table}[th]
\centering
\caption{Hyper-Parameters} 
\begin{tabular}{c|c|c|c|c|c}
\toprule
\hline 
Model & drop prob & batch size & lr & \# of epochs & weight decay\\
\hline
TCN & 0.0527015 & 64 & 0.0011261 & 35 & 5.8373053e-07\\
Deep4Net & 0.5 & 64 & 0.01 & 35 & 0.0005\\
ShallowNet & 0.5 & 64 & 0.000625 & 35 & 0.0\\
EEGNet & 0.25 & 64 & 0.001 & 35 & 0.0\\

\hline 
\bottomrule
\end{tabular} 
\label{tab:hps}
\end{table}

We followed the same experimental setup as in \cite{kiessner2023extended}, which is a recent work on EEG pathology detection using deep learning. To have an unbiased selection of hyper-parameters, We used the hyper-parameters used in this work \cite{kiessner2023extended}. We compared four different neural network architectures: EEGNet, DeepConvNet, ShallowConvNet, and TCN. We trained each model from scratch and fine-tuned a pretrained model on the source dataset. We evaluated each model on both TUAB and NMT datasets using balanced accuracy as the metric.



\newpage
 \bibliographystyle{elsarticle-num-names} 
 \bibliography{ref.bib}





\end{document}